\documentclass[conference]{IEEEtran}
\IEEEoverridecommandlockouts
\usepackage[table,xcdraw]{xcolor}
\usepackage{tabularx,booktabs}
\usepackage{hyperref}
\usepackage{cite}
\usepackage{amsmath,amssymb,amsfonts}
\usepackage{algorithmic}
\usepackage{graphicx}
\usepackage{makecell}
\usepackage{textcomp}
\usepackage{xcolor}
\usepackage{float}
\usepackage{etoolbox}
\usepackage{atbegshi}
\AtBeginDocument{\AtBeginShipoutNext{\AtBeginShipoutDiscard}}
\makeatletter
\patchcmd{\@makecaption}
  {\scshape}
  {}
  {}
  {}
\makeatother
\def\BibTeX{{\rm B\kern-.05em{\sc i\kern-.025em b}\kern-.08em
    T\kern-.1667em\lower.7ex\hbox{E}\kern-.125emX}}

\IEEEpubid{\makebox[\columnwidth]{ \hfill} \hspace{\columnsep}\makebox[\columnwidth]{ }}

\begin{document}

\title{NCL-SM: A Fully Annotated Dataset of Images from Human Skeletal Muscle Biopsies\\

\thanks{EPSRC Centre for Doctoral Training in Cloud Computing for Big Data, Newcastle University}
\thanks {This is a full paper based on the poster and findings presented at ML4H Symposium 2023.}
}\textbf{}

\makeatletter
    \newcommand{\linebreakand}{%
      \end{@IEEEauthorhalign}
      \hfill\mbox{}\par
      \mbox{}\hfill\begin{@IEEEauthorhalign}
    }
    \makeatother

\author{\IEEEauthorblockN{ Atif Khan}
\IEEEauthorblockA{\textit{School of Computing, WCMR} \\
\textit{Newcastle University}\\
Newcastle, UK \\
a.khan21@ncl.ac.uk}
\and
\IEEEauthorblockN{Conor Lawless*}
\IEEEauthorblockA{\textit{WCMR} \\
\textit{Newcastle University}\\
Newcastle, UK \\
conor.lawless@ncl.ac.uk}

\and
\IEEEauthorblockN{Amy E Vincent}
\IEEEauthorblockA{\textit{WCMR} \\
\textit{Newcastle University}\\
Newcastle, UK \\
amy.vincent@ncl.ac.uk}
\and
\linebreakand
\IEEEauthorblockN{Charlotte Warren}
\IEEEauthorblockA{\textit{WCMR} \\
\textit{Newcastle University}\\
Newcastle, UK \\
charlotte.warren@ncl.ac.uk}
{\footnotesize \textsuperscript{*}Corresponding Author}
\and
\IEEEauthorblockN{Valeria Di Leo}
\IEEEauthorblockA{\textit{WCMR} \\
\textit{Newcastle University}\\
Newcastle, UK \\
valeria.di-leo@ncl.ac.uk}
\and
\IEEEauthorblockN{Tiago Gomes}
\IEEEauthorblockA{\textit{WCMR} \\
\textit{Newcastle University}\\
Newcastle, UK \\
tiago.gomes@ncl.ac.uk}
\and
\IEEEauthorblockN{A. Stephen McGough}
\IEEEauthorblockA{\textit{School of Computing} \\
\textit{Newcastle University}\\
Newcastle, UK \\
stephen.mcgough@ncl.ac.uk}

}

\maketitle

\begin{abstract}
Single cell analysis of human skeletal muscle (SM) tissue cross-sections is a fundamental tool for understanding many neuromuscular disorders. For this analysis to be reliable and reproducible, identification of individual fibres within microscopy images (segmentation) of SM tissue should be automatic and precise. Biomedical scientists in this field currently rely on custom tools and general machine learning (ML) models, both followed by labour intensive and subjective manual interventions to fine-tune segmentation. We believe that fully automated, precise, reproducible segmentation is possible by training ML models. However, in this important biomedical domain, there are currently no good quality, publicly available annotated imaging datasets available for ML model training. In this paper we release NCL-SM: a high quality bioimaging dataset of 46 human SM tissue cross-sections from both healthy control subjects and from patients with genetically diagnosed muscle pathology. These images include $>$ 50k manually segmented muscle fibres (myofibres). In addition we also curated high quality myofibre segmentations, annotating reasons for rejecting low quality myofibres and low quality regions in SM tissue images, making these annotations completely ready for downstream analysis. This, we believe, will pave the way for development of a fully automatic pipeline that identifies individual myofibres within images of tissue sections and, in particular, also classifies individual myofibres that are fit for further analysis.

\end{abstract}


\section{Introduction}
\label{sec:intro}
Microscopy imaging and cytometry-based pseudo imaging techniques allow us to observe protein expressions within individual cells within tissue images, when a single cell segmentation approach is used. For many biological and disease processes, this is the most appropriate spatial scale for understanding mechanisms.  Further, the spatial arrangement of cells of different classes within tissues \textit{in vitro} is often informative about biology and disease pathology. 
There are many diseases affecting SM tissue, including amyotrophic lateral sclerosis  \cite{Wales2011SeminarSclerosis}, multiple sclerosis \cite{Filippi2018MultipleSclerosis}, muscular dystrophy \cite{Bushby2010ReviewManagement} and a wide range of mitochondrial diseases \cite{Morrison2009NeuromuscularDiseases}. The dataset we present here is collected from healthy human control subjects and from patients suffering from genetically diagnosed muscle pathology, including mitochondrial diseases. Mitochondrial diseases are individually uncommon but are collectively the most common metabolic disorder affecting 1 in 5,000 people \cite{Ng2016MitochondrialManagement}. They can cause severe disabilities and adversely affect the life expectancy of patients \cite{Barends2016}. Mitochondrial disease pathology is complex and highly heterogeneous \cite{Ng2016MitochondrialManagement}. Some of the latest approaches to classifying mitochondrial diseases, particularly previously unknown mitochondrial diseases, and quantifying disease severity are based on the analysis of single myofibre protein expression profiles and clinical information from patients \cite{Alston2017}.  These same approaches are currently being successfully applied to other types of muscle pathology \cite{DiLeo2023JournalX-xx}.  
\newline Any analysis of SM images at the single myofibre level requires 1) precise segmentation of individual myofibres, as regions close to the myofibre membrane are known to exhibit differential features \cite{Vincent2018SubcellularMuscle}. 2) removal of myofibres damaged by freezing that might occur in the process of storage and thawing, as the subcellular patterns in protein expression, or indeed per-cell mean expression, will be impacted by the technical artefact, masking the target biology. 3) removal of SM myofibres that are not sliced in transverse orientation or are partially sectioned, as the presence of such myofibres does not allow for a standard or uniform comparison across all the myofibres in a tissue and 4) removal of folded tissue. Tissue can fold in on itself during tissue handling and slide preparation. Such folding artificially amplifies apparent protein expression in affected regions and is again a purely technical artefact, not related to target biology.  

Currently most single-myofibre SM analysis is carried out using custom built semi-automatic pipelines like mitocyto \cite{Warren2020}, using general image analysis tools like Ilastik \cite{Berg2019Ilastik:Analysis} or cellprofiler \cite{Carpenter2006CellProfiler:Phenotypes}, or using vanilla ML models like stardist \cite{Schmidt2018} or cellpose \cite{Stringer2020Cellpose:Segmentation}. None of these approaches produce segmentation quality required for analysis of SM out of the box, without suitable training in our experience. Custom built pipelines like mitocyto are used more often than general ML models (which are not trained on SM data) as these require relatively fewer corrections than general ML models. We evaluate and discuss mitocyto segmentation quality in section \ref{Sec:eval}. To improve the segmentation quality and remove compromised myofibres and SM regions, biomedical scientists spend hours manually correcting the issues in a tissue section before doing downstream quantitative analysis. This can be an inefficient use of scientist's time but also the corrections are subjective and not reproducible.

The main barrier to the development of a suitable ML tool/model for fully automatic segmentation and curation of SM myofibres is the lack of any high quality, manually segmented and curated SM myofibre image data on which to train appropriate ML models. In this paper we make available a high quality fully manually segmented and mostly manually curated SM imaging dataset collected using two different imaging technologies: Imaging Mass Cytometry (IMC) and ImmunoFluorescence (IF) amounting to 50,434 myofibres in 46 tissue sections, 30,794 of which are classed as `analysable', 18,102 classed as `not-analysable-due-to-shape', 1,538 myofibres classed as `not-analysable-due-to-freezing-damage' and 405 annotations of folded tissue regions. We describe the Newcastle Skeletal Muscle (NCL-SM) dataset, introduce quantitative quality assessment metrics relevant to single-myofibre SM tissue segmentation and compare annotation quality achieved using our existing pipeline with our new, ground truth manual annotations.
%\newline 

The main contributions of this paper are:
\begin{itemize}
    \item Publishing a high quality, manually segmented set of IMC and IF images of human skeletal muscle tissue, curated by expert biomedical scientists: the NCL-SM dataset. We provide NCL-SM\footnote{\url{https://doi.org/10.25405/data.ncl.24125391}} and our code\footnote{\url{www.github.com/atifkhanncl/NCL-SM} } that evaluates annotation quality and classifies non-transverse sliced myofibres.
    \item Introducing quantitative quality assessment metrics for SM myofibre segmentation. 
    \item Evaluating an existing, semi-automated pipeline applied to NCL-SM and comparing results to clearly highlight where improvements are required and to pave the way for development of an automatic ML pipeline for segmenting single myofibres that are fit for analysis. 
\end{itemize}

\section{Tissue Collection and Image Curation}
\label{sec:review}

The Wellcome Centre for Mitochondrial Research (WCMR)\footnote{\url{www.newcastle-mitochondria.com/}} is one of the leading institutes conducting research into mitochondrial diseases worldwide, and we have an unparalleled repository of clinical data and tissue from healthy control subjects and patients with genetically diagnosed muscle pathology, largely mitochondrial disease patients with mitochondrial myopathy. The NCL-SM dataset includes images of 46 tissue sections that capture spatial variation in protein expression within tissue (including within myofibres). We use microscopy-based techniques i.e. IF and advanced protein expression measurement techniques like IMC that allow us to observe the spatial variation in the expression of up to 40 proteins in tissue simultaneously.

The assessment process before ethical approval for collection of SM tissue from human donors is granted, is rigorous and muscle tissue donations can be painful.  As a result, SM tissue collections are scarce and valuable, requiring available SM tissue sections to be used efficiently. Biomedical scientists in our centre follow processes to preserve and efficiently use the available samples throughout the research cycle. This also applies to the imaging data captured from these tissues i.e. within the images it is critical to efficiently curate all usable myofibres. Analysing all available, usable myofibres is also a way to ensure that no selection bias creeps into the analysis.  We identify all usable/analysable myofibres by precisely segmenting all available SM fibres and curating the subset that are fit for further analysis. Currently, segmentation and curation is a heavily manual process.

Machine learning has made some great strides in image segmentation, object detection and image classification \cite{Minaee2022ImageSurvey, Greenwald2022Whole-cellLearning, Schmidt2018,Stringer2020Cellpose:Segmentation}. However, leveraging ML tools is only possible with large, good quality datasets. 
To our knowledge there is no such publicly available dataset for SM fibre segmentation or classification. Making this dataset available and clearly defining the challenge involved in myofibre segmentation and curating the usable myofibres is a crucial first step towards development of an automatic tool for this problem. 
%\newline

High quality annotated datasets are critical for development of relevant ML models or pipelines. This has been evident since the early days of modern ML with datasets such as MNIST \cite{Deng2012TheResearch}, COCO \cite{Lin2014LNCSContext} and more recently SA-1B \cite{Kirillov2023SegmentAnything} enabling the construction of some seminal ML models like ResNet \cite{He2015DeepRecognition}, VGG \cite{Simonyan}, vision transformer \cite{Dosovitskiy2020ANSCALE} and SAM \cite{Kirillov2023SegmentAnything}.

\section{Capturing images}
The following is the sequential process of collecting SM tissue images.

\begin{table*}[]
    \caption{Annotation counts in NCL-SM dataset. In the table TS, AM, NTM,FAM and FR stands for Tissue Section, Analysable Myofibre, Non-Transverse Myofibre, Freezing Artefact Myofibre and Folded Region respectively. }
    \centering
    \begin{tabular}[width=1\textwidth]{||c c c c c c c||} 
     \hline
     Imaging Technique & TS Count & Myofibre Count & AM Count & NTM Count & FAM Count & FR Count\\  
     \hline\hline
     IMC & 27 & 22,979 & 14,841 & 7,358 & 780 & 84\\ 
     \hline
    IF & 19 & 27,455 & 15,953 & 10,744 & 758 & 321 \\ 
     \hline
     Total & 46 & 50,434 & 30,794 & 18,102 & 1,538 & 405 \\ 
     \hline
    \end{tabular}
    \label{tab:all dataset details}
\end{table*}

\subsection{Biopies}
Following ethical approval from the Newcastle and North Tyneside Local Research Ethics Committee and informed consent from control and patient subjects, biopsies of SM were collected from  patient and healthy control subjects. Specific ethical approval for this project was not required as it consisted of a re-analysis of previously captured image data. This original data was obtained from approved studies of samples from Newcastle Brain Tissue Resource (Approval Ref:2021031) and Newcastle Biobank (Application Ref: 042). 

\subsection{IMC}
Imaging mass cytometry is an imaging technique to simultaneously observe the expression of multiple proteins in a tissue section as described in Figure \ref{Fig:IMC process figure}.

\paragraph{NCL-SM images}
The 46 tissue section images in NCL-SM are made by arranging grayscale images of a cell membrane protein marker i.e. dystrophin and of a mitochondrial mass protein marker VDAC1 into an RGB image where R = membrane protein marker, G =  mass protein marker and B = 0. Each channel of the images are contrast stretched (5 to 95 percentile) to improve image contrast for the segmentation task but raw images without contrast stretching are also included in NCL-SM.

\begin{figure*}[htbp]
\includegraphics[width=1\textwidth]{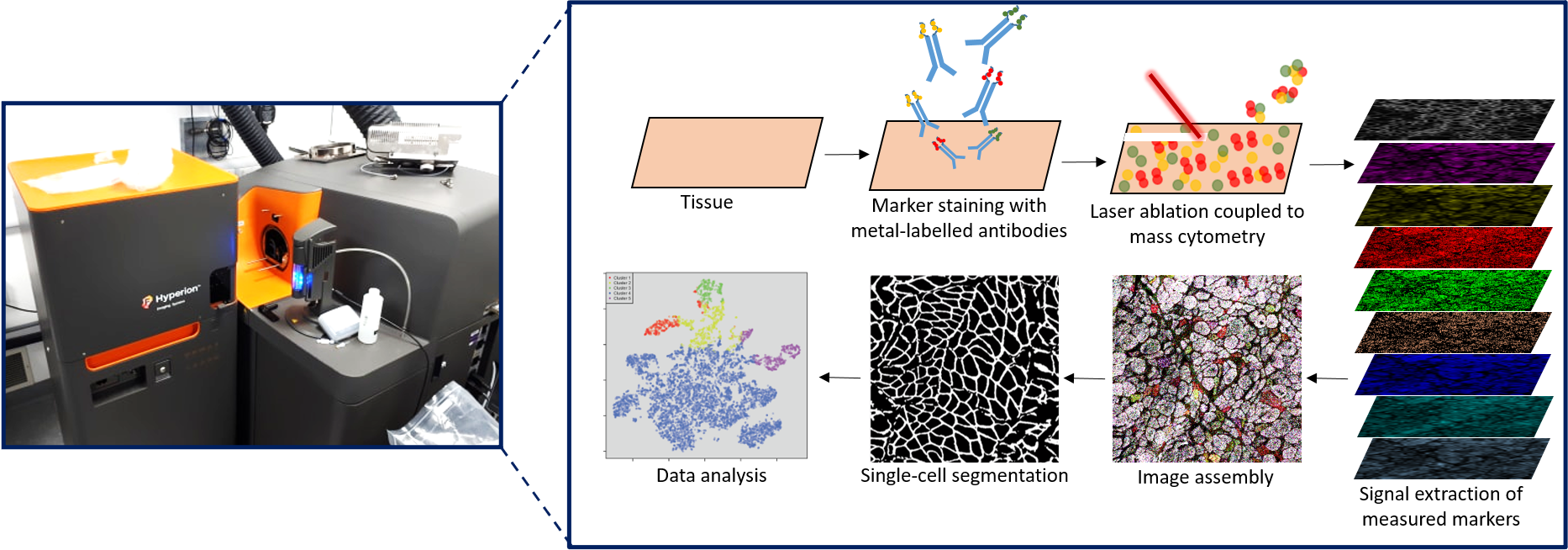}
\caption{\textbf{Imaging Mass Cytometry Experimental Procedure} During IMC myofibres are stained with a panel of antibodies conjugated to heavy metals. Tissue (made up of myofibres) is scanned by a pulsed laser which ablates a spot of tissue section. The tissue is vaporised on each laser shot and enters the mass cytometer where the relative concentration of heavy metals can be quantified. These measurements are later combined as pseudo images, one for each metal/protein, where location and intensity of each pixel corresponds to the amount of metal isotopes at each spot. Myofibres are segmented in these images for conducting single myofibre data analysis. Figure adapted from \cite{Giesen2014HighlyCytometry}}. 
\label{Fig:IMC process figure}
 \end{figure*}

\subsection{IF}
 Immunofluorescence imaging is another technology that uses the antibodies to their antigen to target fluorescent dyes to specific protein targets within a cell/myofibres. It allows the capture of high resolution and high bit-depth microscopy images, but only up to 5 protein targets can be observed simultaneously \cite{Im2019ChapterStaining}.

\subsection{Myofibre Segmentation}
\label{myofibre_segment}
The protocol for myofibre segmentation was i) include all areas within a myofibre that had mitochondrial mass signal, ii) exclude any areas within a myofibre that had myofibre membrane signal and iii) prioritise signal from within myofibre when membrane signal is weak.  Point iii) is necessary because noise is common in these types of data resulting in some overlapping mass and membrane pixels.  In such scenarios, we consider mitochondrial mass signal within a myofibre to be the most reliable indicator of myofibre morphology. 
We employed annotation specialists from Gamaed\footnote{\url{www.gamaed.com/}} working under the close oversight of expert biomedical scientists from WCMR to segment each myofibre in all 46 tissue sections, amounting to 50,434 myofibres. Gamaed annotators used the online Apeer platform\footnote{\url{www.apeer.com/}} for all manual annotations. All segmentation went through rigorous visual inspection and a number of random myofibres in the data were segmented separately by expert biomedical scientists for quality assurance (QA). This is further discussed in Section \ref{Sec: QA}.

\begin{figure}[htbp]
\centerline{\includegraphics[width=.5\textwidth]{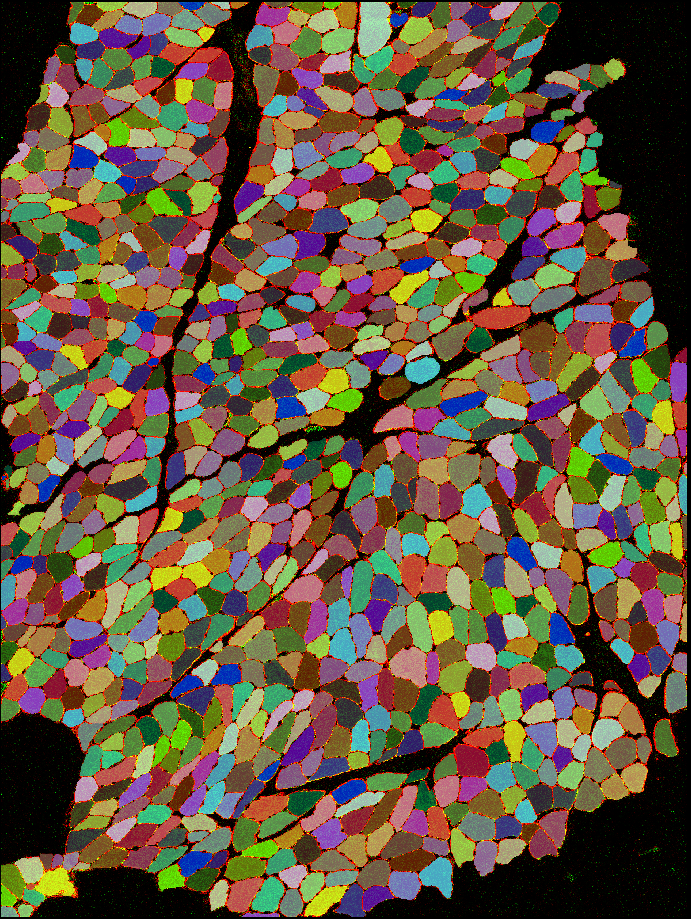}}
\caption{\textbf{Typical Manual Segmentation of a SM Tissue Section}:  An IMC SM tissue section image from subject P17. Consists of 1,068 myofibres manually segmented following the protocol i.e. include all myofibre mass and exclude membrane. }
\label{Fig:Typical myofibre segmentation}
\end{figure}

\subsection{Freezing Artefact Myofibre Classification}
The protocol for identifying myofibres with freezing artefacts was i) look for a leopard spot pattern within myofibres that are typical of freezing damage and ii) look for partial myofibres i.e. large part of myofibres missing as a result of freezing. These patterns are demonstrated in Figure \ref{Fig:Typical freezing damaged fibres}.
%\newline 

The freezing artefact classification annotation was duplicated by two experts from WCMR and any disagreement was resolved by discussion between a panel of experts. This resulted in 1,538 SM fibres with freezing artefacts where both annotators were in agreement.  All annotations which differed between annotators were reviewed and resolved by a second expert after discussion. 

\begin{figure}[htbp]
\centerline{\includegraphics[width=.5\textwidth]{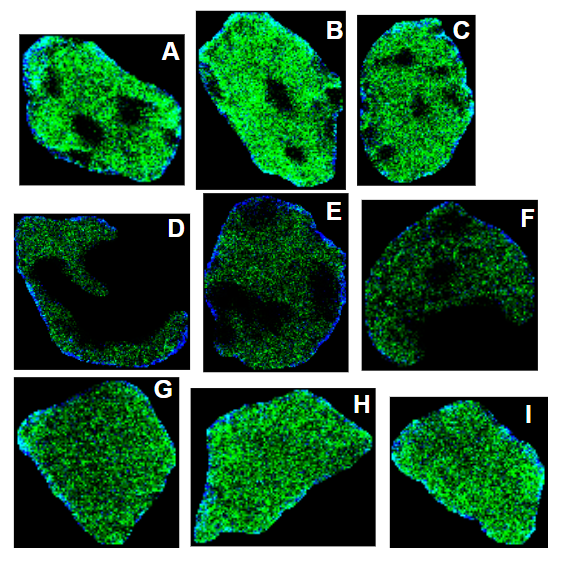}}
\caption{\textbf{Identifying Myofibre Freezing Artefacts} All myofibres from section P02. A (id:190), B (id:138) and C (id:257) are typical freezing damaged myofibres resulting in a leopard spot pattern, D (id:448), E (id:117) and F (id:398) are partial myofibres that are more severely damaged by freezing, G (id:303), H (id:415) and I (id:288) are examples of myofibres without any freezing defects.}
\label{Fig:Typical freezing damaged fibres}
\end{figure}

\subsection{Non-Transverse Sliced Myofibre Classification}
The protocol for identifying non-transverse sliced myofibres was to look for i)  myofibres with skewed aspect ratio e.g. elongated ii) all myofibres at the border of image: these are partial observations iii) segmented objects which are too small or too big and iv) myofibres with unusual convexity. As demonstrated in Figure \ref{Fig:Typical Non-Transverse sliced  fibres},
for annotating non-transverse sliced fibres we employed the following approach 1) two experts working together identified up to 1,500 such myofibres in the data 2) using these 1,500 myofibres, thresholds for area, convexity and aspect ratio were calculated 3) these thresholds and a function to detect any myofibre on the edge were then applied on whole data, resulting in two classes of myofibres. 4) finally both classes of myofibres were rigorously visually inspected to detect and correct any mis-classification. These resulted in 18,102 non-transverse sliced myofibres (NTM).  

\begin{figure}[htbp]
\centerline{\includegraphics[width=.5\textwidth]{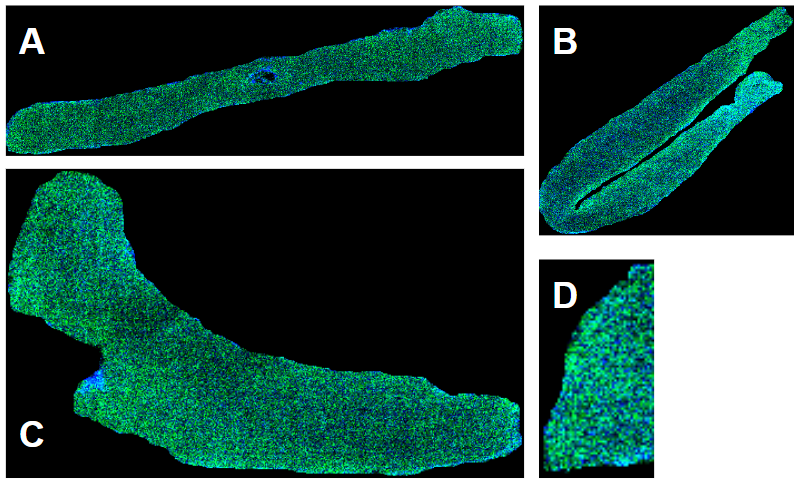}}
\caption{\textbf{Typical Non-Transverse Sliced Myofibres}: Myofibres from tissue section C04. A (ID: 39 ) is a typical elongated myofibre cross-section which has not been cut in the transverse orientation B is myofibre (ID:6) with unusual convexity, `C' is myofibre (ID:24) of large area and `D' is a partial myofibre (ID:12) that has been truncated by the border of the image }
\label{Fig:Typical Non-Transverse sliced  fibres}
\end{figure}

\subsection{Folded Tissue Segmentation}
Folded tissue regions were segmented by an expert biomedical scientist, these were identified by looking for overlapping membrane signal that results in mesh pattern as demonstrated in Figure \ref{Fig:Typical SM Tissue Folding}. This resulted in annotations of 405 different tissue regions affecting 37 out of 46 sections.

\begin{figure}[htbp]
\centerline{\includegraphics[width=.5\textwidth]{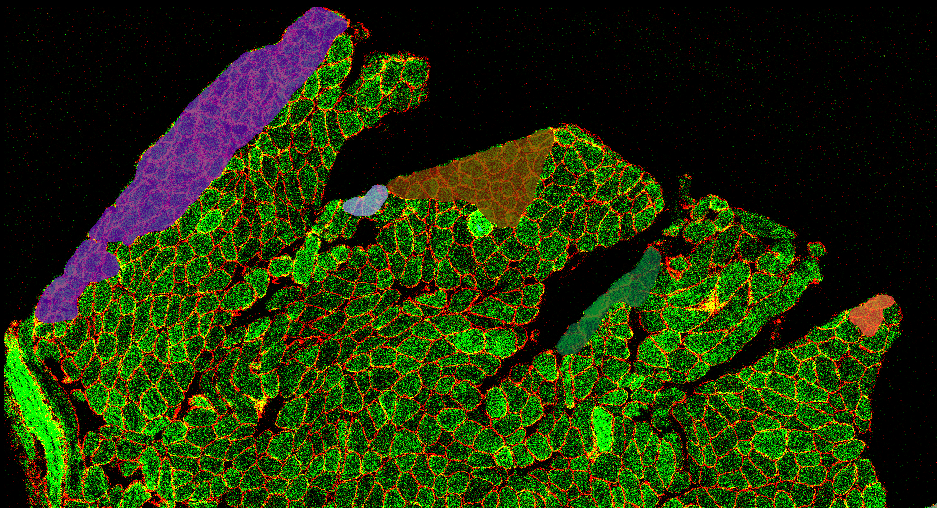}}
\caption{\textbf{Typical SM Tissue Folding}  Segmentation of folded regions in a tissue section P06}
\label{Fig:Typical SM Tissue Folding}
\end{figure}

\section{Quality Evaluation Metrics and Measurements}
\label{Sec: QA}
In order to evaluate the quality of myofibre segmentation annotations we need quantitative metrics that capture the nuances required during SM analysis. In this section we first define these nuances and introduce the quality evaluation metrics, we then present the annotation quality of NCL-SM in terms of these metrics.  
\subsection{Assessment Metrics}

\begin{figure}[htbp]
\centerline{\includegraphics[width=.5\textwidth]{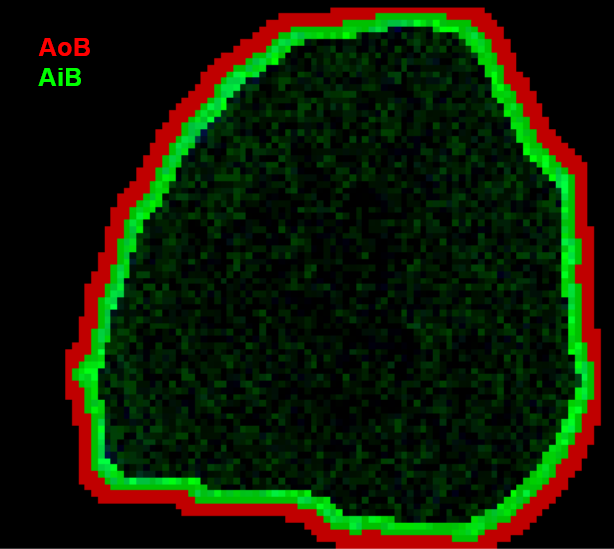}}
\caption{\textbf{Area Near the Membrane}  IMC image of a fibre (ID:723) from tissue section P17 illustrating Area outside the Border (AoB, red) and Area inside the  Border (AiB, green) on either side of the border of an annotated myofibre.  These areas are identified by eroding and dilating the border using a 5x5 and 9x9 pixel kernels for IMC and IF images respectively.}
\label{Fig:Areas_near_myofibre_border}
\end{figure}

Intersection over Union (IoU) is widely used evaluation metric to measure the quality of annotation/segmentation in computer vision tasks. But for myofibre segmentation examining IoU of each myofibre alone will not reveal important aspects about segmentation quality i.e. the area missed or included matter as emphasised in Sections  \ref{sec:intro} and \ref{myofibre_segment}, in other words we want any automatic pipeline to have high accuracy segmenting areas on either side of the border of each myofibre as illustrated in Figure \ref{Fig:Areas_near_myofibre_border}. To measure this, we developed two quantitative metrics: myofibre mass missed correlation ($r_{AoB}$) and myofibre membrane included correlation ($r_{AiB}$)  along with the aforementioned IoU. All three metrics are defined below:

\begin{itemize}

\item \textbf{Myofibre mass missed correlation ($r_{AoB}$) }: This is the Pearson correlation between the proportion of myofibre mass pixels missed in the Area outside the myofibre Border ($AoB$) by annotator $x$, compared to annotator $y$, across all available myofibres $i \in{1...n}$.
$$ \label{AoB}
    \resizebox{0.9\hsize}{!}{%
   $r_{AoB} = \frac{{}\sum_{i=1}^{n} (x_{AoBi} - \overline{x_{AoB}})(y_{AoBi} - \overline{y_{AoB}})}{\sqrt{\sum_{i=1}^{n} (x_{AoBi} - \overline{x_{AoB}})^2  \sum_{i=1}^{n}(y_{AoBi} - \overline{y_{AoB}})^2}}  $%      
        }
$$

where  $x \textsubscript{AoBi}$ is the proportion of myofibre mass pixels in AoB in myofiber $i$ as annotated by annotator $x$; $ y \textsubscript{AoBi}$ is the proportion of myofibre mass pixels in AoB in myofiber $i$ as annotated by annotator $y$; overbar represents the mean across all $n$ myofibres;  and $n$ is the number of myofibres assessed.

\item \textbf{Myofibre membrane included correlation ($r_{AiB}$)}: Defined as above, but for $AiB$:
$$ \label{AiB}
\resizebox{0.9\hsize}{!}{%
   $ r_{AiB} = \frac{{}\sum_{i=1}^{n} (x_{AiBi} - \overline{x_{AiB}})(y_{AiBi} - \overline{y_{AiB}})}{\sqrt{\sum_{i=1}^{n} (x_{AiBi} - \overline{x_{AiB}})^2  \sum_{i=1}^{n}(y_{AiBi} - \overline{y_{AiB}})^2}}   $%      
        } 
$$

where  $x \textsubscript{AiBi}$ is the proportion of myofibre membrane pixels in AiB in myofiber $i$ as annotated by annotator $x$; $ y \textsubscript{AiBi}$ is the proportion of myofibre membrane pixels in AiB in myofiber $i$ as annotated by annotator $y$.

\item \textbf{IoU ($IoU_{i}$)}: This is defined as intersection of overlapping pixels divided by union of all pixels between two annotations of myofibre $i$. This is measured per myofibre and $\overline{IoU}$ is the mean across all $n$ myofibres assessed.

\end{itemize}

\begin{table*}[]
    \caption{ Annotation quality for human-to-human annotation comparison as mention in Section \ref{myofibre_segment}. In the table MF-A, $r \textsubscript{AoB}$, $r \textsubscript{AiB}$, $\overline{IoU}$, ( A\%(IoU $>$0.80), A\%(IoU $>$0.90),  A\%(IoU $>$0.95)), QA-IMC, QA-IF stands for `Myofibres Assessed', `Myofibre Mass Missed Correlation', `Myofibre Membrane Included Correlation', 'Mean IoU', ('Accuracy in terms of \% of myofibres meeting IoU threshold of 0.8, 0.9 and 0.95'), `Quality Assessment for IMC images' and `Quality Assessment for IF images' respectively. }
    \label{tab:benchmark_ann}
    \centering
    \begin{tabular}[width=1\textwidth]{||c c c c c c c c ||} 
    
     \hline
     \newline
     Annotations & MF-A  & $r \textsubscript{AoB}$&  $r \textsubscript{AiB}$ & $\overline{IoU}$ & A\%(IoU $>$0.80) & A\%(IoU $>$0.90) & A\%(IoU $>$0.95) \\  
     \hline\hline
     QA-IMC & 53 & 0.99&  0.77 & 0.96 & 100 & 100& 77.4  \\ 
     \hline
     QA-IF & 23 & 0.92 &  0.94 & 0.96 & 100 & 100& 74  \\ 
     \hline
    
    \end{tabular}
\end{table*}

\begin{table*}[]
     \caption{Annotation quality comparison between NCL-SM (considered as ground truth) and 1) mitocyto+ and 2) QA annotation used as benchmark. In the table TS, mitocyto+ and QA stands for `Tissue sections', `mitocyto with manual corrections' and `QA annotations' respectively. }
      \label{tab:eval_results} 
    \centering
    \begin{tabular}[width=1\textwidth]{||c c c c c c c c c||} 
    
 \hline
 Annotation & TS & MF-A & $r_AoB$&  $r_AiB$  & $\overline{IoU}$ & A\%(IoU$>$0.8) &A\%(IoU$>$0.9) & A\%(IoU$>$0.95)\\ 
 \hline\hline
 mitocyto+ & P02, P06 & 1040 & 0.9 & -0.15 & 0.91 & 95.24 & 74 & 11.3\\ 
 \hline
  QA & P02, P06& 26 & 0.96 & 0.72 & 0.95 & 100 & 100 & 61.5\\ 
 \hline
    \end{tabular}
\end{table*}

\subsection{Assessment Measurements}

As mentioned in Section \ref{myofibre_segment}, for QA during manual segmentation, experts at WCMR routinely annotated random myofibres in various sections to make sure the quality of annotations by specialists from Gamaed remain high quality. The observations are listed in Table \ref{tab:benchmark_ann}. As these are comparison of duplicate manual annotations, we believe this should be used as benchmark for any automatic segmentation tool or pipeline.

\section{Evaluation of Existing Tools on NCL-SM}
\label{Sec:eval}
Considering NCL-SM annotations as `ground truth' we evaluated the segmentation quality produced by mitocyto\footnote{\url{www.github.com/CnrLwlss/mitocyto}} a custom Python image analysis package for semi-automated SM segmentation, built around OpenCV~\cite{Culjak2012AOpenCV}. This package requires manual intervention to correct classifications, which were performed by experts at WCMR. Observations are listed in Table \ref{tab:eval_results}. We can clearly see that for the purpose of analysis of SM at the single myofibre level where precision of segmentation near the membrane of the myofibre is important as described in Section \ref{sec:intro} the current pipeline with manual intervention is not fit for purpose. This is highlighted by our metrics i.e. low values for r\textsubscript{AiB} and accuracy measured in terms of IoU thresholds of 0.9 and 0.95 demonstrate lack of precision segmenting near the membrane.

\section{Discussion}
\subsection{Metrics}
As observed in Section \ref{Sec:eval} the metrics we introduced are capable of quantifying the segmentation quality required for SM analysis at the single myofibre level. It also demonstrates that IoU on its own can be misleading in this case. 
\subsection{Problem definition}
The challenge of segmenting for single myofibre analysis can be divided into two tasks
\begin{itemize}
    \item \textbf{Segmentation}: Each myofibre and folded tissue region needs to be identified in the tissue section image. Following this all folded regions need to be removed and an instance segmentation mask of all remaining myofibres should be made.
    \item \textbf{Classification}: The two types of  not analysable myofibres i.e. `not-analysable-due-to-shape' and `not-analysable-due-to-freezing-damage' need to be identified and removed. Leaving a final instance segmentation mask of curated myofibre that are fit for further analysis.
\end{itemize}

 \subsection{Possible ML Solutions} 
This paper has focused on the collection and curation of the NCL-SM dataset along with defining metrics for analysing the quality of annotation approaches. Although ML approaches for performing segmentation and classification is out of scope for this work we suggest some possible starting points:

\textbf{Workflow}: We believe the sequential workflow that worked in the manual annotation process to create NCL-SM should also work in any automatic pipeline. That is 1) Segment all myofibres in the image to create an instance segmentation mask, 2) segment folded regions in the image to create FR mask, 3) classify NTMs using the myofibres instance segmentation mask to create a NTM mask, 4) use the myofibres instance segmentation mask and the image to create a FAM mask and 5) Subtract masks created in 2,3,4 from 1 to create final instance segmentation mask of 'Analysable Myofibres'.

\textbf{Segmentation}: There is a range of ML models like UNETs \cite{Huang2020}, R-CNNs \cite{Babaie} that might be useful and the metrics we defined above can be used as loss/accuracy metrics for model training.
 \newline \textbf{Classification}: Similarly object detection or image classification models like ResNet \cite{He2015DeepRecognition} and vision transformer \cite{Dosovitskiy2020ANSCALE} might make accurate predictions after training with NCL-SM. 

\section{Conclusion and Future Work}
In this paper we release the NCL-SM dataset, a high quality dataset of $>$ 50k annotated SM myofibre segmentations with a curated subset of myofibres classified as fit for downstream analysis. We describe how we assess segmentation quality for single myofibres, introducing the quantitative metrics relevant for assessment. We demonstrate the high quality of annotation of NCL-SM in terms of relevant metrics. We define the challenges involved in segmenting SM myofibres for possible automatic ML solution. We believe this dataset will enable the development of novel ML solutions to address the problem of automatic segmentation of SM fibres and classification of individual myofibres that are fit for downstream analysis. 
 
NCL-SM consists of $>$ 50k myofibres segmentations but we have significantly fewer FR segmentations i.e. 405 across all images. To address this in our future work 1) we shall expand NCL-SM to include more high quality annotated data not only from our centre but establish a process in place by which others can add their data which meet the quality standards of NCL-SM and 2) we want to to build an automatic ML pipeline to address the challenges highlighted in this paper utilising the NCL-SM dataset.

\section{Acknowledgements}
This work was supported by the EPSRC Centre for Doctoral Training in Cloud Computing for Big Data and Wellcome Centre for Mitochondrial Research. AEV is in receipt of a Newcastle University Academic Track Fellowship. Annotation work was carried out by an independent, specialist data labelling company: \href{https://www.gamaed.com/}{Gamaed}.
\bibliographystyle{ieeetr}
\IEEEtriggeratref{28}
\bibliography{references_use}

\end{document}